\documentclass[icg]{iosart2x}
\usepackage{dcolumn}
\usepackage{float}
\usepackage{multirow}
\usepackage{subfigure}

\begin{document}
\begin{frontmatter} 

\title{Neural Network Structure Design based on N-Gauss Activation Function\thanks{This article was funded by the National Natural Science Foundation of China and National Key Research and Development Program of China(62076028, 2018AAA0101000).}}

\runtitle{Neural Network Structure Design based on N-GAUSS Activation Function}

\author[A,C]{\inits{X.}\fnms{Xiangri} \snm{Lu}\ead[label=e1]{3120195446@bit.edu.cn}},%
\author[A,B]{\inits{H.}\fnms{Hongbin} \snm{Ma}\ead[label=e2]{mathmhb@139.com}}%
\thanks{Corresponding author:Hongbin Ma. \printead{e2}.}
\and
\author[A,C]{\inits{J.}\fnms{Jingcheng} \snm{Zhang}\ead[label=e3]{3220190755@bit.edu.cn}}
\address[A]{State Key Laboratory of Intelligent Decision and Control for Complex Systems, \orgname{Beijing Institute of Technology}, Beijing, \cny{People's Republic of China}\printead[presep={\\}]{e1}}%
\address[B]{Institute of Pattern Recognition and Intelligent Systems, \orgname{School of Automation}, Beijing Institute of Technology, \cny{People's Republic of China}\printead[presep={\\}]{e2}}
\address[C]{Institute of Navigation, Guidance and Control, \orgname{School of Automation}, Beijing Institute of Technology, \cny{People's Republic of China}\printead[presep={\\}]{e3}}

\begin{abstract}
Recent work has shown that the activation function of the convolutional neural network can meet the Lipschitz condition, then the corresponding convolutional neural network structure can be constructed according to the scale of the data set, and the data set can be trained more deeply, more accurately and more effectively. In this article, we have accepted the experimental results and introduced the core block $N-Gauss$, $N-Gauss$, and $Swish$ ($Conv1$, $Conv2$, $FC1$) neural network structure design to train $MNIST$, $CIFAR10$, and $CIFAR100$ respectively. Experiments show that $N-Gauss$ gives full play to the main role of nonlinear modeling of activation functions, so that deep convolutional neural networks have hierarchical nonlinear mapping learning capabilities. At the same time, the training ability of $N-Gauss$ on simple one-dimensional channel small data sets is equivalent to the performance of $ReLU$ and $Swish$.
\end{abstract}

\end{frontmatter}

\section{Introduction}

The activation function plays an important role in understanding the nonlinear structure in the artificial neural network model. In the neuron, the input value is weighted and summed by the activation function to get the activation value. When the activation threshold is reached, the neuron is activated. The activation function is introduced to increase the nonlinearity of the neural network model. If there is no nonlinear activation function, the neural network is equivalent to matrix multiplication. The neural network model includes linear and non-linear functions. The non-linear function is the introduction of the activation function in the neural network.%

In the $1950s$, artificial intelligence was in the connectionist stage, and the neural network in machine learning already had an activation function model. At that time, the activation function was only used as an intermediate link in processing data in the neurons of the neural network model. In the case of limited computing power at that time. Limited by the constraints of computing power, the complexity of the activation function cannot reflect the advantages of the neural network architecture. After $2006$, with the concept of deep belief network proposed by Hitton, the function of the activation function is not only reflected in data processing, but the activation function affects the calculation speed and robustness of the neural network system.%

Common activation functions are Sigmoid function, tanh function, $ReLU$ function and $Softmax$ function. The first three activation functions have been verified by a lot of experiments. The $ReLU$ activation function has better network adaptability than the other two. It is mainly used in the middle layer of neural networks, and $Softmax$ is commonly used in the output layer. In $2017$, the $Swish$ activation function was proposed, and the threshold performance of $Swish$ is smoother than the proposed activation function. In order to ensure the stability of the deep neural network learning process, in the five-layer neural network structure constructed in this article, the first and third layer activation functions both use $ReLU$, $Swish$ or one of the New Gaussian activation functions($N-Gauss$), fully connected layer The activation function of uses one of the above three activation functions, and the output layer uses the $Softmax$ function.

\section{N-GAUSS ACTIVATION FUNCTION}
Since the activation function of the neural network is forward-trained, the training data needs to be backpropagated for the training of the neural network, which requires the activation function to have the basic properties of the derivative function in most of the interval, that is, the function is in the derivable interval Meet Lipschitz conditions.By assuming that if the function $f(x)=tanh(x)$ satisfies the existence of $k>0$, any $\mathrm{x}, \mathrm{y} \in \operatorname{dom}(\mathrm{f})$ has $|f(x)-f(y)| \leqslant k|x-y|$.

\subsection{Problem raised}
\emph{Question: Can $f(x)/x$ satisfy the Lipschitz condition?}

\emph{Answer: If the function $f(x)=tanh(x)$ satisfies the existence of $k>0$ so that any $\mathrm{x}, \mathrm{y} \in \operatorname{dom}(\mathrm{f})$ has $|f(x)-f(y)| \leqslant k|x-y|$, then Let $f(x)$ be the Lipschitz function. If $f(x)$ is the Lipschitz function defined on $(-\infty,-1] \cup[1,+\infty)$, then f(x)/x is also the Lipschitz function defined on $(-\infty,-1] \cup[1,+\infty)$.}

\subsection{Proof of the problem}
\emph{Proof: If you want to prove that the function $f(x)/|x|$ satisfies the Lipschitz condition on the interval $(-\infty,-1] \cup[1,+\infty)$, the pre-proof $[1,+\infty)$ holds. Since $f(x)$ is a function defined on $[1,+\infty)$, so $f(x)$/x is also a function defined on $[1,+\infty)$. Since $f(x)$ is a Lipschitz function defined on $[1,+\infty)$, there is $k>0$ such that any $x, y \in[1,+\infty)$ has $|f(x)-f(y)| \leqslant k|x-y|$.
Now prove that $f(x)/x$ is bounded on $[1,+\infty)$.}

$\because \mathrm{x} \in[1,+\infty)$,

$\therefore|\mathrm{f}(\mathrm{x})-\mathrm{f}(1)| \leqslant \mathrm{k}|\mathrm{x}-1|$

$\because|\mathrm{f}(\mathrm{x})|-|\mathrm{f}(1)| \leqslant|\mathrm{f}(\mathrm{x})-\mathrm{f}(1)|$ and $|\mathrm{x}-1| \leqslant|\mathrm{x}|+1$,

$\therefore|\mathrm{f}(\mathrm{x})| \leqslant|\mathrm{x}|+1+|\mathrm{f}(1)|$,

$\because \mathrm{x} \geqslant 1$,

$\therefore|\mathrm{f}(\mathrm{x}) / \mathrm{x}| \leqslant 1+(1+|\mathrm{f}(1)|) / \mathrm{x} \leqslant 2+|\mathrm{f}(1)|$,

That is, $\mathrm f\mathrm(x) / \mathrm(x)$ is bounded on $(1,+\infty)$.

And

$\because \mathrm{x}, \mathrm{y} \in[1,+\infty)$,

$\therefore|f(x) / x-f(y) / y|$

$=|(y f(x)-x f(y)) / x y|$

$=|y f(x)-y f(y)+y f(y)-x f(y) / x y|$

$\leqslant|f(x)-f(y) / x|+|f(y)(y-x) / x y|$

$\leqslant k|x-y|+|f(y) / y||x-y|$

$\leqslant(k+2+|f(1)|)|x-y|$,

So $f(x)/x$ is the Lipschitz function defined on $[1,+\infty)$,$f(x)/x$ is also a Lipschitz function defined on $(-\infty,-1]$.
In summary, $f(x)/x$ is the Lipschitz function defined on $(-\infty,-1] \cup[1,+\infty)$.

According to this, the N-Gauss type activation function can be used as the activation function in the interval $(-\infty,-1] \cup[1,+\infty)$. The function image is shown in $Figure 1$.
\begin{figure}[H]
\includegraphics[width=0.5\textwidth]{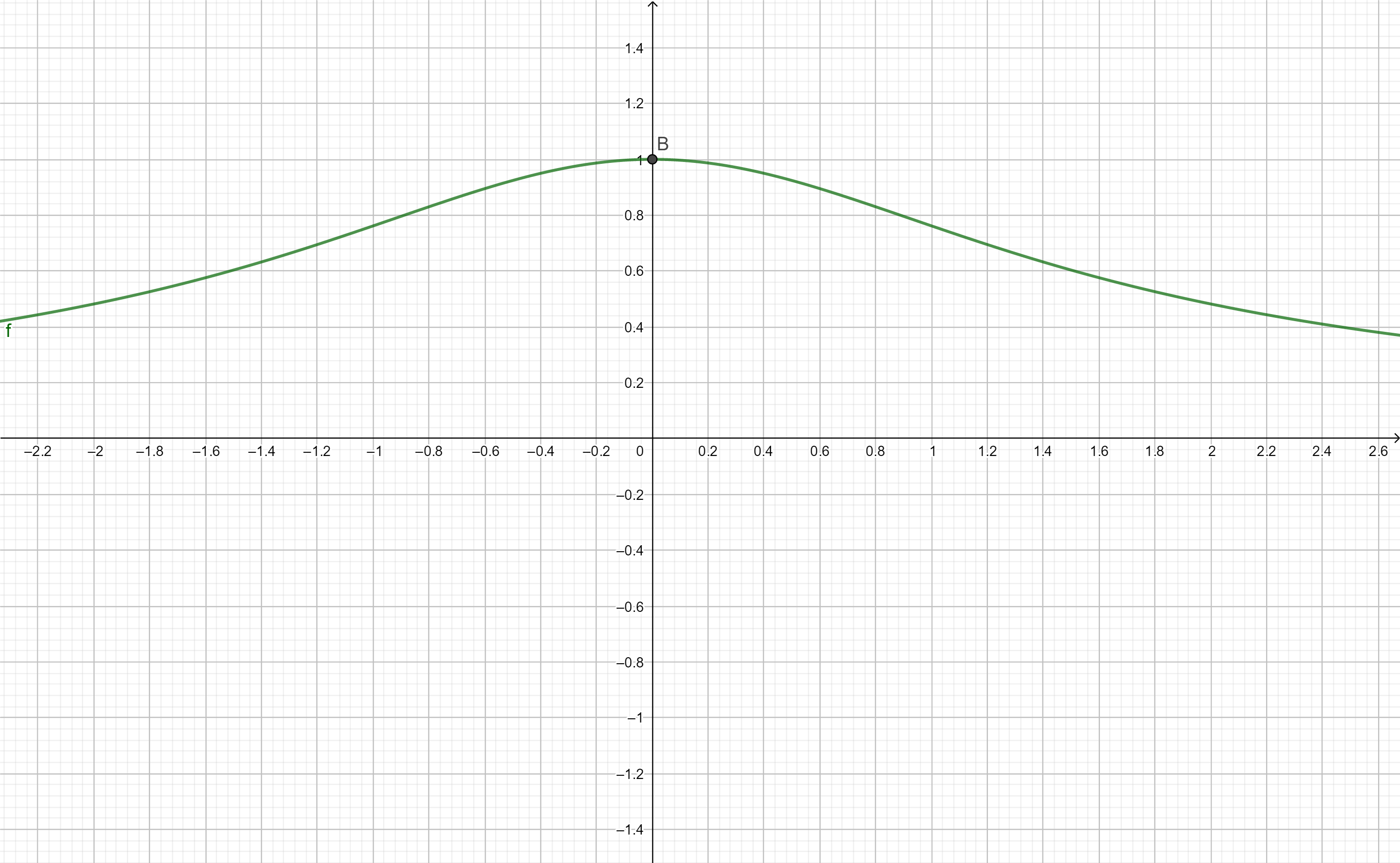}
\caption{N-Gauss activation function image.}
\end{figure}

\section{NEURAL NETWORK ARCHITECTURE DESIGN}
An important part of the neural network is the construction of the network structure. The experiment uses two convolutional layers, a pooling layer, a fully connected layer and an output layer, as shown in $Figure 2$.
\begin{figure}[H]
\includegraphics[width=0.95\textwidth]{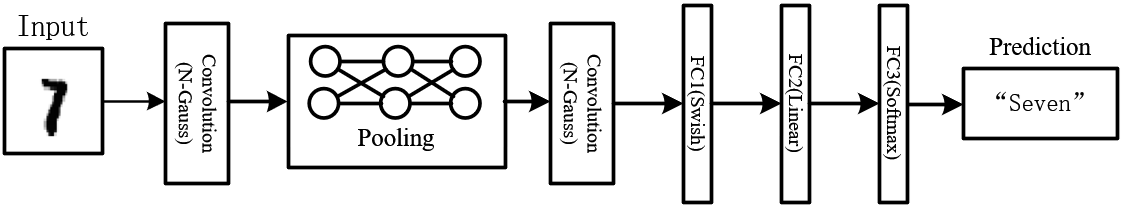}
\caption{Schematic diagram of convolutional neural network NNS structure.}
\end{figure}

The first convolutional layer of the convolutional neural network shown in the figure above defines the input channel as 1, the output channel as 10, and the size of the convolution kernel is $5*5$; then the maximum pooling operation is performed on the convolution result of the first layer,the pooling size of the transformation layer is $2*2$, and the step size is 2. The third convolutional layer defines the input channel as 10, the output channel as 20, and the size of the convolution kernel is $3*3$; the input scale of the first fully connected layer is 2000, and the number of neurons of output is 500; the input scale of the second fully connected layer is 500, and the number of output neurons is 10, that is, the digital image $MNIST$ and $CIFAR10$ data sets are divided into ten categories; if it is the $CIFAR100$ data set, then The number of neurons of output is adjusted to 100, that is, it is divided into 100 categories. The detailed data is shown in $Table 1$.
\begin{table}[H]
\caption{Design of Convolutional Neural Network Architecture for MNIST and CIFAR Data Sets} \label{t1}
\resizebox{\textwidth}{!}{
\begin{tabular}{|c|l|l|l|l|l|l|c|c|c|}
\hline
\multirow{2}{*}{Layers} & \multicolumn{3}{c|}{Input Size}                                                          & \multicolumn{3}{c|}{Output Size}                                                      & \multicolumn{3}{c|}{Move Window Size}                                \\ \cline{2-10}
                        & MNIST                      & CIFAR10                      & CIFAR100                     & MNIST                      & CIFAR10                    & CIFAR100                    & MNIST & \multicolumn{1}{l|}{CIFAR10} & \multicolumn{1}{l|}{CIFAR100} \\ \hline
Conv1                   & \multicolumn{1}{c|}{28*28} & \multicolumn{1}{c|}{3*32*32} & \multicolumn{1}{c|}{3*32*32} & 10*24*24                   & 30*28*28                   & 300*28*28                   & 5*5   & 5*5                          & 5*5                           \\ \hline
Pooling                 & 10*24*24                   & 30*28*28                     & 100*28*28                    & 10*12*12                   & 30*14*14                   & 300*14*14                   & 2*2   & 2*2                          & 2*2                           \\ \hline
Conv2                   & 10*12*12                   & 30*14*14                     & 100*14*14                    & 20*10*10                   & 60*12*12                   & 600*12*12                   & 3*3   & 3*3                          & 3*3                           \\ \hline
FC1                     & 20*10*10                   & 60*12*12                     & 600*12*12                    & \multicolumn{1}{c|}{500*1} & \multicolumn{1}{c|}{500*3} & \multicolumn{1}{c|}{5000*3} & \multicolumn{3}{c|}{fully-connected, Linear or Non-linear}                         \\ \hline
FC2                     & \multicolumn{1}{c|}{500*1} & \multicolumn{1}{c|}{500*3}   & \multicolumn{1}{c|}{5000*3}  & \multicolumn{1}{c|}{10*1}  & \multicolumn{1}{c|}{10*3}  & \multicolumn{1}{c|}{100*3}  & \multicolumn{3}{c|}{fully-connected, Linear}                         \\ \hline
FC3                     & --                         & --                           & --                           & \multicolumn{1}{c|}{10*1}  & \multicolumn{1}{c|}{10*1}  & \multicolumn{1}{c|}{100*1}  & \multicolumn{3}{c|}{fully-connected, Softmax}                        \\ \hline
\end{tabular}}
\end{table}

The design of neural network architecture mainly includes the design of the intermediate hidden layer architecture of the neural network and the matching problem of the activation function in it. Now we will discuss how to solve the problem of activation function collocation between hidden layers. According to the introduction, the activation function of the hidden layer in the convolutional neural network will use the single and double structure combination of the $N-Gauss$ function, $ReLU$ and $Swish$ function for architectural design (Odd and even). Now number the above functions: $N-Gauss$, $ReLU$, $Swish$. Then the structure of the convolutional neural network has the following 9 schemes:$NNN$,$RRN$,$SSN$,$NNR$,$RRR$,$SSR$,$NNS$,$RRS$,$SSS$.

\section{experiment}
This experiment is on the hardware platform $Intel(R)$ $Core(TM)$ $i7-10750H$ $CPU$ $@2.60GHz$ and $NVIDIA$ $GeForce$ $GTX$ $1660 Ti$ $GPU$; the software platform is $jupyter notebook$ and $Python3.7$ programming language.
We use hardware to prove that the $N-Gauss$ activation function can be used in neural network training on several benchmark data sets, and explore the $N-Gauss$ activation function of the forward convolution layer in the built convolutional neural network combined with the fully connected layer with the effectiveness of the $Swish$ activation function neural network structure.The experimental data of the loss value of the data set under the structure of the convolutional neural network reorganization activation function are shown in $Table 2$.
\begin{table}[H]
\caption{Convolutional neural network reorganizes the loss value of the data set under the structure of the activation function} \label{t1}
\begin{tabular}{|c|c|c|c|c|c|}
\hline
\multirow{2}{*}{Conv1} & \multirow{2}{*}{Conv2} & \multirow{2}{*}{FC1}     & \multicolumn{3}{c|}{Loss rates on datasets} \\ \cline{4-6}
                       &                        &                          & MNIST        & CIFAR10      & CIFAR100      \\ \hline
N-Gauss                & N-Gauss                & \multirow{3}{*}{N-Gauss} & --           & --            & --            \\ \cline{1-2} \cline{4-6}
ReLU                   & ReLU                   &                          & --            & --            & --             \\ \cline{1-2} \cline{4-6}
Swish                  & Swish                  &                          & --            & --           & --             \\ \hline
N-Gauss                & N-Gauss                & \multirow{3}{*}{ReLU}    & 0.0453       & 0.8857       & --             \\ \cline{1-2} \cline{4-6}
ReLU                   & ReLU                   &                          & 0.0328       & 0.4158       & 0.8726        \\ \cline{1-2} \cline{4-6}
Swish                  & Swish                  &                          & 0.0327       & 0.3731       & --             \\ \hline
N-Gauss                & N-Gauss                & \multirow{3}{*}{Swish}   & 0.0482       & 0.3581       & 0.8429        \\ \cline{1-2} \cline{4-6}
ReLU                   & ReLU                   &                          & 0.0323       & 0.4054       & 0.8947        \\ \cline{1-2} \cline{4-6}
Swish                  & Swish                  &                          & 0.0324       & 0.3751       & 0.8522        \\ \hline
\end{tabular}
\end{table}

\subsection{DATASET}
MNIST. The data set is a handwritten digit database created by Google Labs and the Courant Institute of New York University. Its training database has $60,000$ handwritten digital images, and the test database has $1,000$ digital images. The size of each image is $28*28$. The training library $train-images.idx3-ubyte$, the test library $t10k-images.idx3-ubyte$.

CIFAR. $CIFAR10$ data set has a total of $60,000$ color images, each with a size of $32*32$, divided into 10 categories, each with 6000 images. There are 50,000 images used for training, forming 5 training batches, each batch of 10,000 images; the other 10,000 are used for testing, forming a single batch. In the data of the test batch, it is taken from each of the 10 categories, and 1000 sheets are randomly selected from each category. Note that the number of images in a training batch is not necessarily the same. Looking at the training batch in general, there are 5000 images in each category. The $CIFAR100$ data set has 100 classes, and each class contains 600 images. Among the 600 images, there are 500 training images and 100 test images. The 100 categories are actually composed of 20 categories (each category contains 5 subcategories).
\subsection{DATA ANALYSIS}
From the table data, it can be concluded that the activation function $N-Gauss$ is used in the fully connected layer $FC1$ of the neural network, and the three activation functions of $N-Gauss$, $ReLU$ and $Swish$ are used in the convolutional layer. The training does not converge, indicating that the $N-Gauss$ function is not suitable for data classification in the fully connected layer; when $ReLU$ is used as the activation function of the fully connected layer, $N-Gauss$ and $Swish$ are used as the activation function of the convolutional layer, and the $CIFAR100$ training data does not converge. $N-Gauss$ and $Swish$ in the function image is a non-monotonic activation function. When learning a data set, the neural network will have positive and negative features, while $ReLU$ is a positive activation function. The training of the data set will continue in a momentum direction, resulting in too many types of $CIFAR100$ data and causing non-convergence in training; From the data table, the fully connected layer($FC1$) is more robust for the activation function of $Swish$,other structures same as what making $N-Gauss$, $N-Gauss$, and $Swish$($Conv1$, $Conv2$, $FC1$) of the activation function of the neural network structure in the basically training results.
\begin{figure}[htbp]
\centering
\subfigure[Loss value based on NNS convolutional neural network architecture.]{
\includegraphics[width=5.5cm]{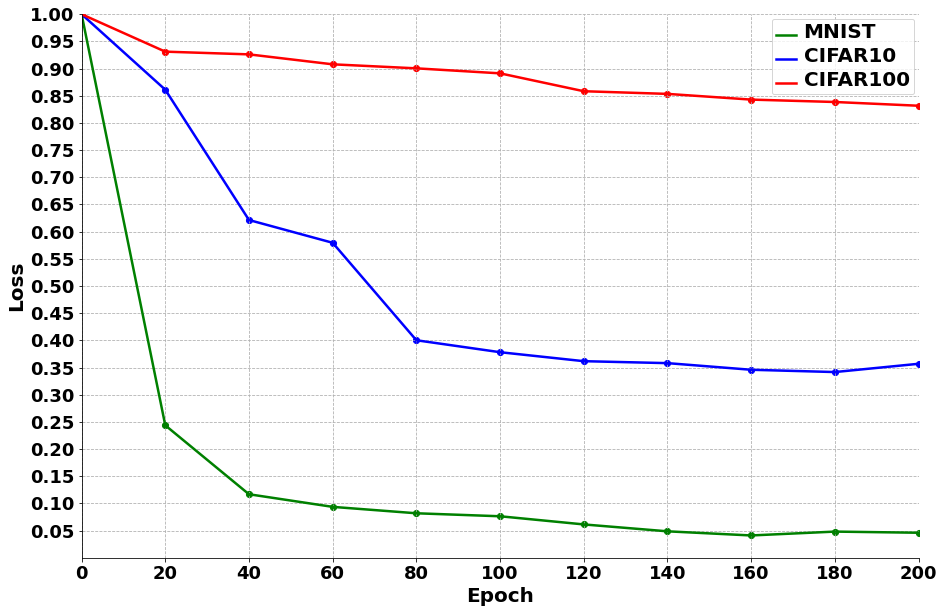}
}
\quad
\subfigure[Loss value based on SSS convolutional neural network architecture.]{
\includegraphics[width=5.5cm]{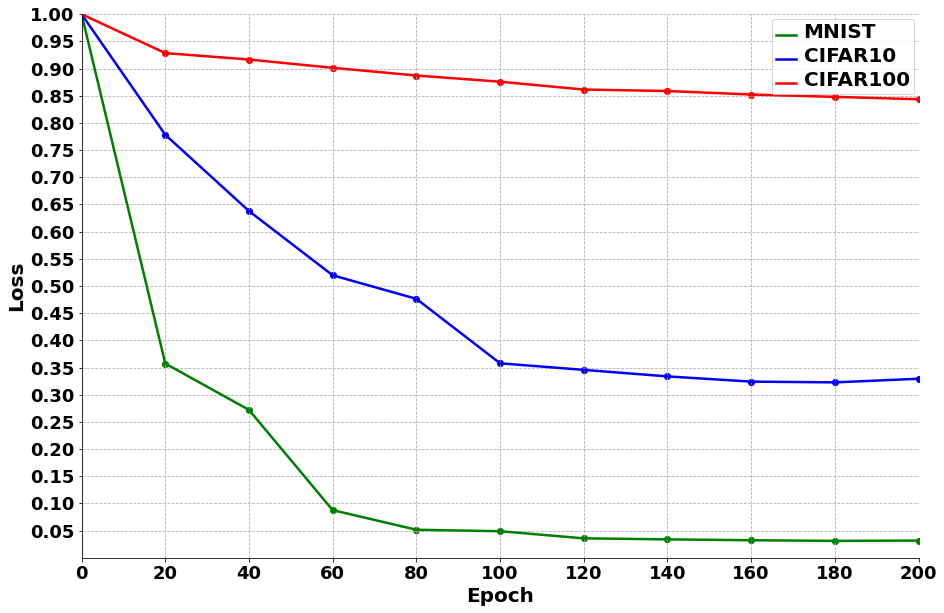}
}
\quad
\subfigure[Loss value based on RRR convolutional neural network architecture.]{
\includegraphics[width=5.5cm]{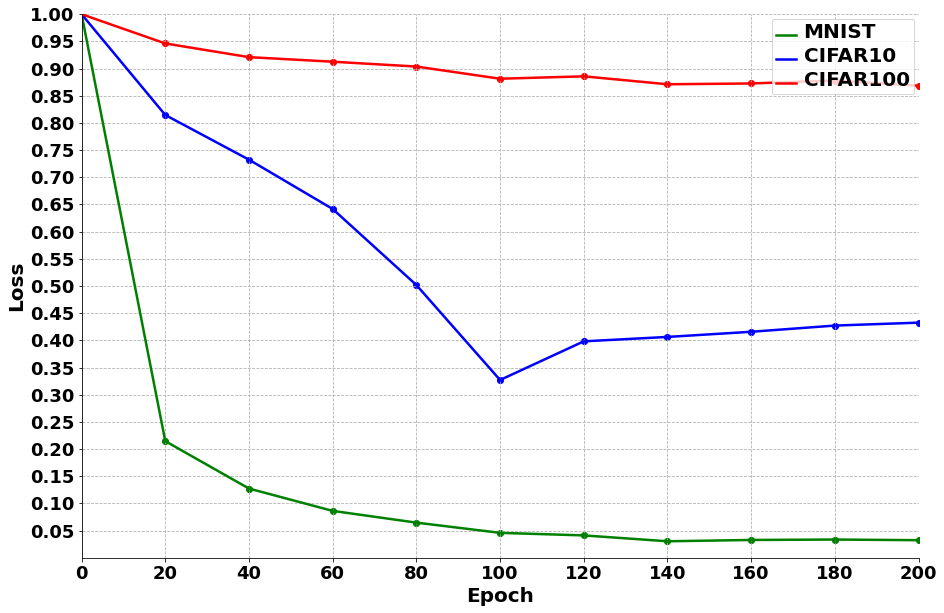}
}
\quad
\subfigure[Loss value based on RRS convolutional neural network architecture.]{
\includegraphics[width=5.5cm]{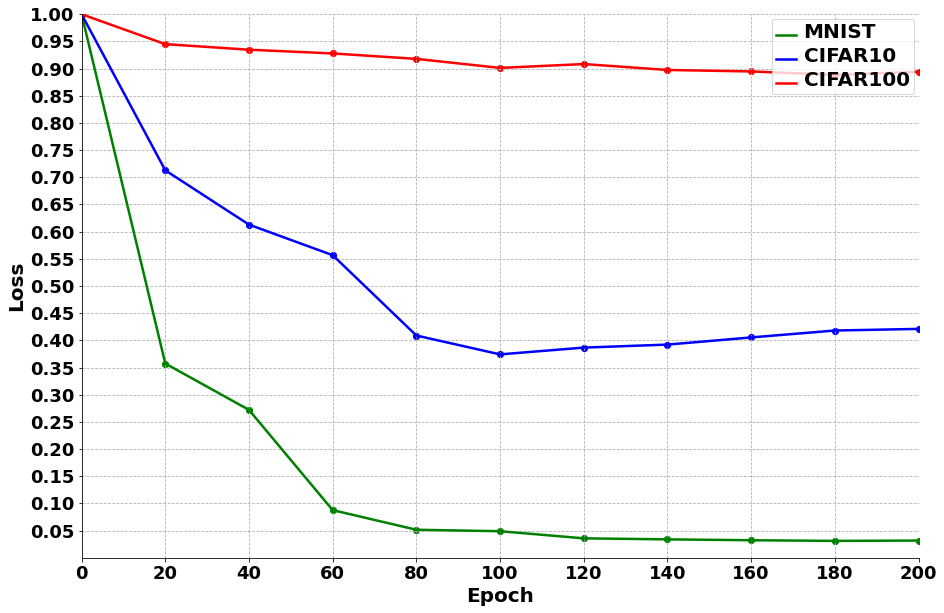}
}
\caption{The loss values generated by different convolutional neural network architectures when training the MNIST, CIFAR10 and CIFAR100 datasets.}
\end{figure}

$Figure 3$ shows the loss value of the convolutional neural network architecture formed by the combination of different activation functions of the hidden layers of the convolutional neural network($Conv1$,$Conv2$,$FC1$)when training the $MNIST$,$CIFAR10$ and $CIFAR100$ datasets. $Figures (a)$ and $(b)$respectively show the training loss values under the $NNS$ and $SSS$ convolutional structures. The results in the figure show the loss values generated by these two convolutional architectures when training the $MNIST$,$CIFAR10$ and $CIFAR100$ data sets,the size and changing trend are basically the same. That is to say, the effect of the convolutional layer with the $N-Gauss$ function as the activation function is basically equivalent to the $Swish$ activation function ($Swish$ is called the best activation function). The stable period of the $NNS$ convolutional neural network structure training $MNIST$ and $CIFAR10$ data sets appears in the $80$ period. In terms of training speed, $NNS$ is faster than the $RRR$ and $RRS$ stable periods ($MNIST:120$ and $CIFAR10:100$). The loss of $NNS$ in the $MNIST$ and $CIFAR10$ data sets stabilized between $(0.030-0.050)$ and $(0.350-0.400)$, respectively. Due to design experiments to verify the effectiveness of some convolutional neural networks, the system hyper parameters are not adjusted. As a result, the loss value monitoring of $CIFAR100$ for the four convolutional structures shows that they are all at $0.85$. The $NNS$ convolution structure is slightly less robust to a large variety of data sets ($CIFAR100$), which is also an area of improvement in the later convolutional neural network structure.
\section{Conclusion}
This article designs and demonstrates the feasibility of the $N-Gauss$ activation function. From the side experiments, it proves that the activation function in the $NNS$ structure is feasible and provides a method to find the activation function, that is, the activation function used. Need to meet $Lipschitz$ conditions. The theory proves the applicability of the activation function and also needs to cooperate with the design of neural grid structure to converge the data training. In this paper, the neural network structure shown in $Figure 2$ is designed, in which the core blocks $N-Gauss$, $N-Gauss$, and $Swish$($Conv1$, $Conv2$, $FC1$) are designed to basically meet the test requirements of small data sets, which shows that $N-Gauss$ activation function can be used for data training and testing, but the loss value of the large-scale data set system is still very high. The reason is the structure design of the neural network. The channel and size of each data set are different. Structural design needs further exploration.
\begin{acks}
This research is partially supported by Nature Science Foundation of China with ID 62076028 and by National Key Research and Development Program of China with ID 2018AAA0101000.
\end{acks}

\end{document}